%% file: main.tex
\documentclass[letterpaper]{article} 
\pdfoutput=1
\usepackage[final,nonatbib]{neurips}
 
\usepackage{cite}
\usepackage{amsmath,amssymb,amsfonts}
\usepackage{algorithmic}
\usepackage{graphicx}
\usepackage{textcomp}
\usepackage{xcolor}
\usepackage{makecell}
\usepackage{setspace}

\usepackage{times}
\usepackage{latexsym}
\usepackage[T1]{fontenc}
\usepackage[utf8]{inputenc}
\usepackage{microtype}
\usepackage{inconsolata}

\usepackage{tabularx}
\usepackage[most]{tcolorbox}
\usepackage{hyperref}       
\usepackage{url}            
\usepackage{booktabs}       
\usepackage{amsfonts}       
\usepackage{bbm}
\usepackage{nicefrac}       
\usepackage{amsmath}
\usepackage{ulem}
\usepackage{multirow}
\usepackage{array}
\newcolumntype{C}[1]{>{\centering\arraybackslash}p{#1}}

\usepackage{float}
\usepackage{threeparttable}
\usepackage{graphicx}
\usepackage{quoting}
\usepackage{xspace}
\usepackage{caption}
\usepackage{subcaption}
\usepackage[dvipsnames]{xcolor}
\usepackage{bm}
\usepackage{longtable}

\usepackage[most]{tcolorbox}
\tcbset{
    mybox/.style={
        colbacktitle=red!50!white,
        colback=red!10!white,    
        colframe=red!30!white,   
        fonttitle=\bfseries\large, 
        fontupper=\ttfamily\normalsize, 
        before upper=\setstretch{1.2},
        boxrule=1mm,      
        title=#1            
    }
}

\newcommand{\redtt}[1]{\textcolor{red}{{\uline{#1}}}}

\def\BibTeX{{\rm B\kern-.05em{\sc i\kern-.025em b}\kern-.08em
    T\kern-.1667em\lower.7ex\hbox{E}\kern-.125emX}}
\begin{document}

\title{Iterative LLM-Based Generation and Refinement of Distracting Conditions in Math Word Problems}

\input{assets/author}

\maketitle

\begin{abstract}
\input{0abstract}
\end{abstract}

\input{1intro}

\input{2relatedwork}

\input{3setup}

\input{4result}

\input{5analysis}

\input{6closing}

\clearpage
\bibliographystyle{splncs04}
\bibliography{ref}
\end{document}

%% file: assets/author.tex
\author{
Kaiqi Yang$^{1}$, Hang Li$^{1}$, Yucheng Chu$^{1}$, 
Zitao Liu$^{2}$, Mi Tian$^{3}$, Hui Liu$^{1}$,\\
$^{1}$ Michigan State University, East Lansing, USA\\
$^{2}$ Jinan University, Guangzhou, China\\
$^{3}$ TAL Education Group, Beijing, China\\
\texttt{\{kqyang, lihang4, chuyuch2, liuhui7\}@msu.edu},\\
\texttt{liuzitao@jnu.edu.cn, tianmi@tal.com}
}

%% file: 0abstract.tex
Mathematical reasoning serves as a crucial testbed for the intelligence of large language models (LLMs), and math word problems (MWPs) are a popular type of math problems. Most MWP datasets consist of problems containing only the necessary information, while problems with distracting and excessive conditions are often overlooked. Prior works have tested popular LLMs and found a dramatic performance drop in the presence of distracting conditions. However, datasets of MWPs with distracting conditions are limited, and most suffer from lower levels of difficulty and out-of-context expressions. This makes distracting conditions easy to identify and exclude, thus reducing the credibility of benchmarking on them. Moreover, when adding distracting conditions, the reasoning and answers may also change, requiring intensive labor to check and write the solutions. To address these issues, we design an iterative framework to generate distracting conditions using LLMs. We develop a set of prompts to revise MWPs from different perspectives and cognitive levels, encouraging the generation of distracting conditions as well as suggestions for further revision. Another advantage is the shared solutions between original and revised problems: we explicitly guide the LLMs to generate distracting conditions that do not alter the original solutions, thus avoiding the need to generate new solutions. This framework is efficient and easy to deploy, reducing the overhead of generating MWPs with distracting conditions while maintaining data quality.

%% file: 1intro.tex
\section{Introduction~\label{sec:intro}}
Large Language Models (LLMs) have become popular tools for supporting teaching and learning~\cite{Luo2023WizardMathEM,Imani2023MathPrompterMR}. Among various educational tasks, solving math word problems (MWPs) is a core component of STEM education~\cite{Cartwright2022WhatsTD,CopurGencturk2021StrategicCF}. A math word problem is a problem presented in a scholastic context that requires applying mathematical operations to structured numerical information to derive a solution~\cite{verschaffel2020word}. To solve MWPs, students must understand the context, extract relevant quantities and relationships, reason from the given information, and finally infer the answer~\cite{Wang2016CognitiveAL}. Given these reasoning requirements, MWPs have become a popular testbed for LLMs to evaluate their mathematical and logical reasoning abilities~\cite{cobbe2021training,bubeck2023sparks}. Recent studies~\cite{Wei2022ChainOT,Zhou2022LeasttoMostPE,Luo2023WizardMathEM} have focused on leveraging LLMs’ capabilities in mathematical operations and logical reasoning, benchmarking and improving their performance on such tasks. However, most existing MWPs contain only the information necessary for solving the problems, while those with irrelevant or distracting conditions remain scarce. Educational and psychological studies have shown that MWPs with distracting or irrelevant information require more advanced cognitive functions to solve~\cite{Hoyer1979EffectsOV,Pasolunghi1999WorkingMA,reusser1988problem}, posing additional challenges for mathematical training.

Although LLMs demonstrate impressive performance in solving and explaining MWPs from elementary to undergraduate levels, irrelevant conditions remain a critical challenge for their reasoning. When irrelevant information is present, LLMs tend to incorporate it into their reasoning processes, often leading to redundant or even incorrect solutions. Taking GSM-8K~\cite{cobbe2021training} as the base datasets, GSM-Symbolic~\cite{Mirzadeh2024GSMSymbolicUT} evaluates state-of-the-art LLMs on MWP datasets both with and without irrelevant conditions, revealing significant performance drops across models of different scales. GSM-8K-Adv~\cite{anantheswaran2024investigating} introduces a pipeline that generates irrelevant conditions for MWPs via adversarial augmentation, forming a loop that alternates between generation and solving, and trains a fine-tuned model on the resulting adversarial instances. GSM-IC~\cite{Shi2023LargeLM} adopts a template-based approach to produce irrelevant conditions, where pre-selected topics (e.g., \textit{age or height of a human}) are converted into templates and inserted into existing MWPs. Although these works generate readable and fluent conditions, the resulting problems are often of low quality. As illustrated in Table~\ref{mwps_with_irrelevant}, GSM-8K-Adv includes examples with entirely off-topic conditions (e.g., \textit{humidity level of house}), which are unrelated to any entities in the original problem. Since other relevant conditions focus on the same topic (\textit{number of cats}), the added condition is obviously irrelevant and easily ignored. Similarly, in GSM-IC, although the question and conditions concern ages, the added irrelevant condition introduces a new character (\textit{Claire's father}). It is evident that the question is about Jessica, and only Claire—not her father—is directly related to the target relationships. Furthermore, most existing works~\cite{Mirzadeh2024GSMSymbolicUT,Shi2023LargeLM} rely on a predefined set of templates in which entities and attributes are chosen based on the creators’ intuition. This not only leads to off-topic content, as observed in GSM-8K-Adv, but also makes the datasets dull and rigid: once students or models recognize the patterns (e.g., sentences about \textit{environmental humidity} are always irrelevant), the intended challenge of such problems diminishes.

\input{assets/examples}

To address these issues, we propose \textbf{IGC-MWP} (Iterative Generation and Checking for Math Word Problems), an iterative framework that uses LLMs to generate distracting conditions for MWPs. We impose no constraints on the topics or entities used in distracting conditions; instead, we design a set of prompts that guide LLMs to generate, reflect on, and revise the added content. The process follows a top-down sequence: a distracting condition is first elicited through a generation prompt defining the task and basic rubrics. Then, two sets of prompts are used to evaluate the difficulty and quantitative consistency of the generated condition, ensuring nontrivial results. Subsequently, two additional steps assess the desirable and undesirable characteristics of the newly created MWPs, further improving the added conditions and the overall problem quality. From the second step onward, LLMs are asked to judge candidate conditions and determine whether they satisfy the requirements specified in the current prompt by providing verdicts and explanations. If a verdict is negative, the problems, candidate conditions, and verdict–explanation pairs are fed back to the previous step. The rubrics are then augmented with the reasons for failure, and the LLMs are prompted to regenerate. This iterative loop allows LLMs to automatically reflect on and refine their generations step by step. In summary, the contributions of this work are as follows:

\begin{itemize}
    \item We propose a novel framework for generating distracting, irrelevant conditions for MWPs.
    \item We design a five-step prompt set that guides LLMs to generate and refine distracting conditions automatically.
\end{itemize}

%% file: assets/examples.tex
\begin{table*}[h]
\centering\normalsize
\begin{tabular}{C{0.2\linewidth} | p{0.7\linewidth}}
\toprule

\textbf{Dataset} &
\multicolumn{1}{p{0.7\linewidth}}{\centering \textbf{Example Math Word Problem (MWP)}} \\ 

\midrule

\textbf{GSM-8K-Adv~\cite{anantheswaran2024investigating}} &
A pet store had 38 siamese cats and 25 house cats, \redtt{with a humidity level of 50\%}. During a sale they sold 45 cats, \redtt{and the humidity level rose to 70\%}. \textbf{\textless\textless{}Question\textgreater\textgreater{}} How many cats do they have left? \\

\midrule

\textbf{IC-GSM~\cite{Shi2023LargeLM}} &
Jessica is six years older than Claire. In two years, Claire will be 20 years old. \redtt{Twenty years ago, the age of Claire’s father is 3 times of Jessica’s age}. \textbf{\textless\textless{}Question\textgreater\textgreater{}} How old is Jessica now? \\

\midrule

\textbf{IGC-MWP (Ours)} &
Albert is wondering how much pizza he can eat in one day. He buys 2 large pizzas and 2 small pizzas. A large pizza has 16 slices, a small pizza has 8 slices, \redtt{and a medium pizza has 12 slices}. If he eats all the large and small pizzas, how many pieces does he eat that day? \\

\bottomrule
\end{tabular}
\vspace{5pt}
\caption{Examples of MWPs with distracting (irrelevant) conditions from GSM-8K-Adv~\cite{anantheswaran2024investigating}, IC-GSM~\cite{Shi2023LargeLM}, and IGC-MWP (ours). The added irrelevant conditions are marked with \redtt{underlines}.}
\label{mwps_with_irrelevant}
\end{table*}

%% file: 2relatedwork.tex
\section{Related Work}
Large language models (LLMs) have demonstrated remarkable performance in mathematical reasoning, particularly when equipped with curated prompting strategies~\cite{Patel2021AreNM,Ahn2024LargeLM,Macina2023MathDialAD}. The Chain-of-Thought (CoT) approach~\cite{Wei2022ChainOT} instructs LLMs to reason step by step, substantially improving accuracy across various reasoning tasks. Least-to-Most (LTM) prompting~\cite{Zhou2022LeasttoMostPE} decomposes complex problems into simpler subproblems, thereby facilitating progressive reasoning. Self-consistency~\cite{wang2022self} encourages LLMs to generate multiple reasoning paths and determine the final answer through majority voting. ReAct~\cite{Yao2022ReActSR} combines reasoning with external tool usage, enabling LLMs to retrieve necessary information and enhance accuracy.

Beyond prompting, several studies have improved LLMs through fine-tuning or instruction optimization. WizardMath~\cite{Luo2023WizardMathEM} proposes an Evol-Instruct method that performs both upward and downward evolution—creating harder and easier problems—to expand a model’s mathematical capability across difficulty levels. MAmmoTH~\cite{Yue2023MAmmoTHBM} constructs \textit{MathInstruct}, an instruction-tuning dataset integrating 13 popular math datasets, and jointly trains LLMs on both textual and code-based reasoning tasks. Despite these advances, most evaluations have been conducted on math word problems (MWPs) that contain only essential information~\cite{Anantheswaran2024CuttingTT}, while MWPs with irrelevant or distracting conditions remain largely unexplored.

Recent studies have shown that such distracting conditions can significantly degrade LLM reasoning. GSM-Symbolic~\cite{Mirzadeh2024GSMSymbolicUT} generates variants of MWPs by first extracting symbolic templates and then refilling them with new variables. GSM-8K-Adv~\cite{anantheswaran2024investigating} adopts an adversarial framework that alternates between generating and solving MWPs to improve both data quality and model robustness. GSM-IC~\cite{Shi2023LargeLM} creates irrelevant conditions using predefined templates—such as a person’s age, height, or shoe size—and inserts them into existing MWPs. While these methods effectively augment MWPs with additional conditions, they depend heavily on fixed human-written templates and thus exhibit limited diversity and topic flexibility.

In contrast to these template-driven approaches, our framework imposes no constraints on topics or entities. Instead, it guides LLMs through general rubrics and iterative reflection, enabling more flexible, context-aware, and creative generation of distracting conditions that better mimic realistic educational settings.

%% file: 3setup.tex
\section{Framework of IGC-MWP}
\subsection{Pipeline Overview}

We design a novel framework, \textbf{IGC-MWP}, for generating distracting conditions in math word problems (MWPs). The framework leverages the reasoning and generation capabilities of large language models (LLMs) to reduce intensive manual labor. All components in this pipeline are implemented using LLMs. Specifically, the model is prompted to add a distracting condition to a given MWP, and then to verify the correctness and quality of the resulting problem. Using the reasoning explanations produced during these checking steps, the LLM further improves the generated problems by identifying deficiencies and regenerating the distracting conditions as needed. The overall framework is illustrated in Figure~\ref{fig:framework}.

\newcommand{\colorparallelogram}{%
  \tikz[baseline]{
    \draw[fill=yellow, draw=YellowOrange, thick]
      (0,0) -- (0.6,0) -- (0.8,0.3) -- (0.2,0.3) -- cycle;
    }%
}
\newcommand{\colordiamond}{%
  \tikz[baseline]{
    \draw[fill=yellow, draw=YellowOrange, thick]
        (-0.3,0.15) -- (0,0.3) -- (0.3,0.15) -- (0,0) -- cycle;
    }%
}

\newcommand{\colorboxshape}[2]{%
  \tikz[baseline]{
    \draw[fill=#1, draw=#2, thick]
        (0,0) -- (0,0.3) -- (0.6,0.3) -- (0.6,0) -- cycle;
    }%
}

\newcommand{\colorfoldershape}[2]{%
  \tikz[baseline]{
    \draw[fill=#1, draw=#2, thick]
        (0,0) -- (0,0.3) -- (0.4,0.3) -- (0.42,0.32) -- (0.58,0.32) -- (0.6,0.3) -- (0.6,0) -- cycle;
    }%
}

\begin{figure*}[h]
    \centering
    \includegraphics[width=0.99\textwidth]{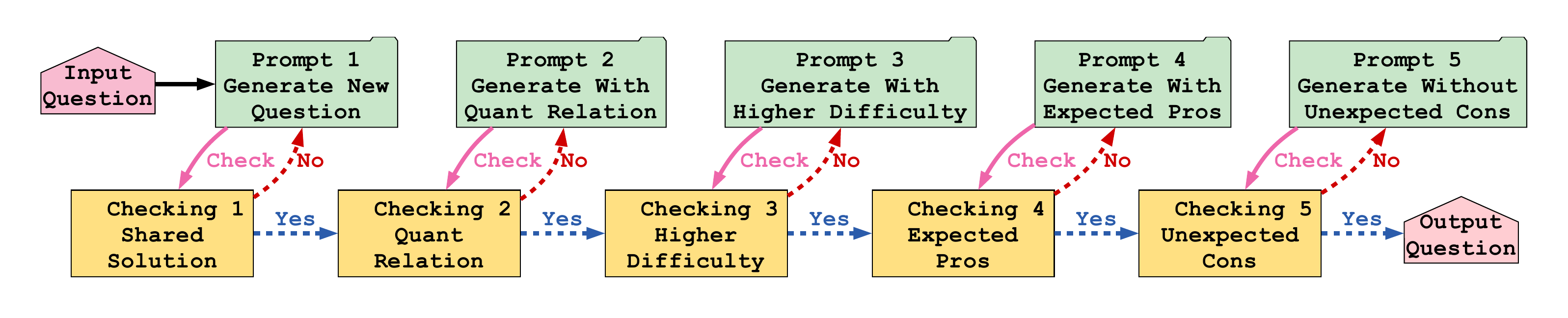} 
    \caption{Framework of our irrelevant condition (IC) generation. 
    The \textcolor{YellowOrange}{Yellow \protect\colorboxshape{yellow}{YellowOrange}} rectangles indicate the judgment mechanisms, while the \textcolor{ForestGreen}{Green \protect\colorfoldershape{YellowGreen}{ForestGreen}} folders represent the prompts along with their generation and revision mechanisms. Dashed arrows denote the flow of judgment outputs (\textcolor{blue}{Yes: $\dashrightarrow$} and \textcolor{red}{No: $\dashrightarrow$}), while solid arrows denote the transmission of textual content.}
    \label{fig:framework}
\end{figure*}

A key challenge in problem generation is the heavy demand for data annotation~\cite{Tan2024LargeLM}. MWPs are typically solved with paragraph-length explanations that describe the relationships among entities along with corresponding formulas or computations~\cite{Wei2025ASO}. To eliminate human annotation or additional solution generation, our method constrains the new MWP (with an irrelevant condition) to share the same solution as the original. This design ensures consistent solution quality while significantly reducing annotation effort.

\subsection{Prompt Design~\label{promptdesign}}

To generate distracting conditions for given MWPs, we design a set of prompts for both generation and validation. Drawing on educational and cognitive psychology research~\cite{sweller1994cognitive,newell1958elements}, we recognize that creating a valid additional condition that satisfies multiple requirements is cognitively demanding. Therefore, we decompose this complex task into several manageable sub-problems, each focusing on a specific and consistent goal, thereby reducing the cognitive load on LLMs. For our task, we split the generation of irrelevant conditions into five hierarchical sub-tasks and design five corresponding prompts.

\subsubsection{Initial Generation~\label{sec:generate1}}

The first sub-task involves the generic generation of distracting conditions, defined as:

\noindent\textit{\textbf{Step 1:}} The LLM is prompted to add one redundant condition while keeping the original solution unchanged. At this stage, no specific rules constrain the model’s creativity; only basic rubrics such as “add one condition,” “maintain the same solution,” and “avoid introducing new background information” are provided. Unlike prior works~\cite{Shi2023LargeLM,anantheswaran2024investigating}, which combine complex and detailed requirements into a single step, our approach isolates this stage to focus solely on generation, thereby improving both clarity and efficiency.

To further enhance task comprehension, we adopt the few-shot methodology~\cite{brown2020language,chowdhery2023palm}, incorporating one demonstration of an MWP with an added irrelevant condition in each prompt. Examples are shown in Table~\ref{tab:demonstrations}.

\input{assets/demostration}

For each MWP, the original problem and its shared solution are provided to the LLM as a pair, and a revised version is generated. An example prompt is shown in Figure~\ref{originalprompt}.
\input{assets/prompt1}

\subsubsection{Quantitative Relationship and Difficulty Checking}
After the initial generation step, we introduce two subsequent stages to evaluate and refine the revised problems at a coarse-grained level. Ideally, the added conditions should establish quantitative relationships with the entities in the original context~\cite{Cartwright2019ExecutiveFI}. However, LLMs often generate trivial cases: (1) purely narrative statements without numerical relationships (e.g., \textit{"his bike is blue and gray"}), or (2) quantitative but non-challenging details that simplify reasoning (e.g., \textit{"he asks for three each year"} when the original problem already defines a clear pattern). Although these outputs technically satisfy the irrelevance requirement, they fail to introduce meaningful cognitive distraction.

\noindent\textit{\textbf{Step 2:}} This step checks whether the newly added condition maintains valid quantitative relationships with the original problem entities, ensuring that it is numerically grounded rather than merely descriptive.

\noindent\textit{\textbf{Step 3:}} This step evaluates whether the condition increases the reasoning difficulty, preventing trivial augmentations that oversimplify or accidentally reveal key information. If a condition fails either check, the LLM is instructed to analyze its deficiencies and regenerate the condition accordingly. The feedback explanations are fed back to the previous step for iterative refinement. Detailed prompt designs for these two checks are provided in Figure~\ref{prompt2} and Figure~\ref{prompt3}.

\input{assets/prompt2}
\input{assets/prompt3}

\subsubsection{Desirable and Undesirable Characteristic Checking}

For problems that pass both the quantitative and difficulty checks, we further apply finer-grained rubrics to refine subtler aspects of quality. Building on prior research in MWP design, we develop prompts that target both desirable and undesirable characteristics to ensure contextual coherence and instructional effectiveness.

\noindent\textit{\textbf{Step 4:}} Desirable traits include maintaining contextual consistency, preserving logical relationships among existing entities, and ensuring that the added condition naturally aligns with the original narrative.

\noindent\textit{\textbf{Step 5:}} Undesirable traits include introducing new or unrelated story elements, creating ambiguous pronoun references, or explicitly signaling irrelevance (e.g., \textit{"but he doesn't buy any"} or \textit{"but she is not interested in it"}). Detailed prompt examples for these two checks are shown in Figure~\ref{prompt4} and Figure~\ref{prompt5}.

\input{assets/prompt4}
\input{assets/prompt5}

\subsection{Rejection Mechanism~\label{sec:regeneration}}

At each step $k$, another LLM evaluates whether the generated distracting condition meets the current criteria. The model produces a verdict along with short reasoning explaining its decision. If the output fails, the intermediate result is rejected and sent back for revision, combining the prior prompt with the new explanations for regeneration. If it passes, the problem proceeds to step $k{+}1$. After the final undesirable-trait check, all successfully validated problems are included in the IGC-MWP dataset. The prompt is shown in Figure~\ref{prompt6}. 

\input{assets/prompt6}

%% file: assets/demostration.tex
\begin{table*}[h]
\centering\normalsize
\begin{tabular}{C{0.2\linewidth} | p{0.7\linewidth}}
\toprule
\textbf{Revised Problems} &
\multicolumn{1}{p{0.7\linewidth}}{\centering \textbf{Example Math Word Problem (MWP)}} \\ 
\midrule

\textbf{Problem 1} & 
Mark has a garden with flowers. He planted plants of three different colors in it. Ten of them are yellow, and there are 80\% more in purple. There are only 25\% as many green flowers as there are yellow and purple flowers. \redtt{Additionally, the number of trees is half the number of flowers.} How many flowers does Mark have in his garden? \\

\midrule

\textbf{Problem 2} & 
Tina makes \$18.00 an hour. If she works more than 8 hours per shift, she is eligible for overtime, paid at 1.5 times her hourly wage. \redtt{If she works less than 4 hours per shift, \$2 per shift is deducted for transportation.} If she works 10 hours per day for 5 days, how much money does she make? \\

\midrule

\textbf{Problem 3} & 
Weng earns \$12 an hour for babysitting, \redtt{and earns \$5 for every 30 minutes of housecleaning.} Yesterday, she only did 50 minutes of babysitting. How much did she earn? \\

\bottomrule
\end{tabular}
\vspace{5pt}
\caption{Examples of MWPs with irrelevant conditions in our work. The added irrelevant conditions are marked with \redtt{underlines}.}
\label{tab:demonstrations}
\end{table*}

%% file: assets/prompt1.tex
\begin{figure*}[!ht]
\begin{tcolorbox}[mybox={Initial Generation Prompts}]
Revise this question by adding one irrelevant condition or value, while keeping the solution still the same. 

The irrelevant condition or value do not alter the reasoning path of the question. 

The irrelevant condition or value does not introduce new background stories, rather, it is related to the entities in original question. 

Directly give the full revised question without explanations. 

\medskip

\textbf{Example}

\textbf{Original Question:} {Five friends eat at a fast-food chain and order the following: 5 pieces of hamburger that cost $\$3$ each ... How much will each of them pay if they will split the bill equally?}

\textbf{Shared Solution:} The cost of 5 pieces of hamburger is \$3 x 5 = \$<<3*5=15>>15 ... So their total bill is \$15 + \$4.80 + \$2.50 +\$2.7 = <<15+4.8+2.5+2.7=25>>25. Hence, each of the five friends will contribute \$25/5 = \$<<25/5=5>>5.

\#\#\#\# 5

\textbf{Added Irrelevant Condition:} \redtt{There will be additional $\$5$ service fee when the total bill is more than $\$26$.}

\textbf{New Question:} {Five friends eat at a fast-food chain and order the following: 5 pieces of hamburger that cost $\$3$ each ... \redtt{There will be additional $\$5$ service fee when the total bill is more than $\$26$}. How much will each of them pay if they will split the bill equally?}

\medskip

\textbf{Original Question:} \textit{\{Question\}}

\textbf{Shared Solution:} \textit{\{Solution\}}

\textbf{New Question:}
\medskip
\end{tcolorbox}

\caption{Example of Step 1 Prompts \textit{Initial Generation}. The added irrelevant conditions are marked with \redtt{underlines}.}
\label{originalprompt}
\end{figure*}

%% file: assets/prompt2.tex
\begin{figure}[!ht]
\begin{tcolorbox}[mybox={Quantitative Relationship Checking Prompts}]
Compare the two questions, find the condition that Q2 have but Q1 does not have. Check whether this condition introduces any new quantitative relationships or new values. If so, just say YES without other explanation; if not, explain why (within 100 words). Start your response with "YES" or "NO".

\textbf{Q1:} \textit{\{Original Question\}}

\textbf{Q2:} \textit{\{Revised Question\}}

\end{tcolorbox}

\caption{Example of Step 2 Prompts \textit{Quantitative Relationship Checking}}
\label{prompt2}
\end{figure}

%% file: assets/prompt3.tex
\begin{figure}[!ht]
\begin{tcolorbox}[mybox={Difficulty Level Checking Prompts}]
These questions have the same solution; check if Q2 is more challenging to solve than Q1 by adding irrelevant conditions. Possible cases include: Q2 introduces seemingly relevant but useless conditions; Q2 introduces seemingly relevant values useless in solution; Q2 decomposes conditions or values with additional solution steps. If Q2 is more challenging than Q1, just say YES without other explanation; if not or similar, explain why (within 100 words). Start your response with "YES" or "NO".

\textbf{Q1:} \textit{\{Original Question\}}

\textbf{Q2:} \textit{\{Revised Question\}}

\end{tcolorbox}

\caption{Example of Step 3 Prompts \textit{Difficulty Level Checking}}
\label{prompt3}
\end{figure}

%% file: assets/prompt4.tex
\begin{figure*}[!ht]
\begin{tcolorbox}[mybox={Desirable Characteristics Checking Prompts}]
Compare the two questions, find the additional condition in Q2 compared with Q1. Check if the additional condition complies with all the following rules one by one. If it complies with all rules, just say YES; if not, explain why (within 100 words). Start your response with "YES" or "NO".

\textbf{Q1:} \textit{\{Original Question\}}

\textbf{Q2:} \textit{\{Revised Question\}}

Rules: The additional condition 
\begin{itemize}
    \item should be counterparts or features of existing entities in Q1; 
    \item should have direct quantitative relationships with the entities in Q1; 
    \item should mention something seems related to the story but useless for solving the Q2; 
    \item should be consistent and reconcilable with the story in Q1; 
    \item should avoid conjunctions of contrast (e.g., however, but, though); 
    \item should keep the range covered by plural pronouns the same.
\end{itemize}
\end{tcolorbox}

\caption{Example of Step 4 Prompts \textit{Desirable Characteristics Checking}}
\label{prompt4}
\end{figure*}

%% file: assets/prompt5.tex
\begin{figure*}[!ht]
\begin{tcolorbox}[mybox={Undesirable Characteristics Checking Prompts}]
Compare the two questions, find the additional condition in Q2 compared with Q1. Check if the additional condition complies with all the following rules one by one. If it complies with all rules, just say YES; if not, explain why (within 100 words). Start your response with "YES" or "NO".

\textbf{Q1:} \textit{\{Original Question\}}

\textbf{Q2:} \textit{\{Revised Question\}}

Rules: The additional condition 
\begin{itemize}
    \item should NOT alter the reasoning path of the original question; 
    \item should NOT introduce new background stories; 
    \item should NOT be impossible to lead to any other feasible calculation-needed solution even the question is changed; 
    \item should NOT make the pronouns ambiguous; 
    \item should NOT give explicit hints stating the additional condition is unnecessary to solve the question (e.g., 'but he doesn't buy any', 'but she is not interested in it', 'but it isn't counted in');
\end{itemize}
\end{tcolorbox}

\caption{Example of Step 5 Prompts \textit{Undesirable Characteristics Checking}}
\label{prompt5}
\end{figure*}

%% file: assets/prompt6.tex
\begin{figure*}[!ht]
\begin{tcolorbox}[mybox={Regeneration Prompts}]
The question you generate does not comply with all the requirements. Considering the following explanation on your errors, revise the question again. Take the error and initial requirements into consideration. Directly give the full revised question without explanations.

\textbf{Q1:} \textit{\{Original Question\}}

\textbf{Q2:} \textit{\{Revised Question\}}

\textbf{Explanation:} The solution does not fit the question because the equation used in the solution is incorrect ... should be 5, not 6.

\end{tcolorbox}

\caption{Example of Regeneration Prompts}
\label{prompt6}
\end{figure*}

%% file: 4result.tex
\section{Experiment and Result}

\subsection{Settings}

In this work, we select GSM-8K as the base dataset. GSM-8K is a benchmark dataset of elementary-level math word problems (MWPs), containing 7.5K training samples and 1K test samples. Each problem typically requires 2 to 8 reasoning steps and involves only basic arithmetic operations ($+, -, \times, \div$). We do not distinguish problems by their number of steps or operation types. Both the generation and evaluation processes in our framework are powered by GPT-3.5-turbo as the backbone LLM.

\subsection{Generated Dataset~\label{result1}}

We first present qualitative examples from our proposed IGC-MWP dataset. Figure~\ref{fig:example1} shows several representative problems generated by our framework.

\input{assets/example1}

As illustrated, the problems in the IGC-MWP dataset exhibit clear advantages and avoid the undesirable characteristics discussed earlier. For instance, in Problem 1 of Figure~\ref{fig:example1}, the original story describes slices of pizza, and the added condition naturally follows this theme rather than introducing unrelated context. The original question involves two pizza sizes, while the added condition introduces a medium pizza with a comparable number of slices (12). This new condition adds meaningful distraction—students must now reason about the number of slices in each pizza type and correctly distinguish between large, small, and medium sizes.  

Similarly, in Problems 2 and 3 (Figure~\ref{fig:example1}), the added conditions involve the number of lines on pages or the number of pages in books, both of which are conceptually tied to the core entities of the original stories about reading. These examples demonstrate that our generated distracting conditions maintain contextual coherence while effectively increasing reasoning difficulty.

\subsection{Performance Drop with the IGC-MWP Dataset~\label{result2}}

In addition to the qualitative analysis in Sec.~\ref{result1}, we quantitatively evaluate LLM performance on the IGC-MWP dataset. Intuitively, well-designed distracting conditions should make problems more challenging, resulting in a greater performance gap relative to the clean dataset.  

For comparison, we include two existing GSM-8K variants: GSM-8K-Adv and GSM-Symbolic, both of which augment the original dataset with distracting conditions. As shown in Table~\ref{table1}, all GSM variants cause notable performance drops compared to the base GSM-8K. Importantly, the performance degradation on our IGC-MWP dataset is the largest among them, suggesting that our generated problems are of higher quality and present more realistic and cognitively demanding distractions for LLMs.

\input{assets/table1}

%% file: assets/example1.tex
\begin{figure*}[!t]
\begin{tcolorbox}[mybox={Examples of IGC-MWP Dataset}]

\textbf{Problem 1:} Albert is wondering how much pizza he can eat in one day. He buys 2 large pizzas and 2 small pizzas. A large pizza has 16 slices, a small pizza has 8 slices, \redtt{and a medium pizza has 12 slices}. If he eats all the large and small pizzas, how many pieces does he eat that day?

\textbf{Problem 2:} Maila is reading a 120-page book. Yesterday, she was able to read 12 pages and today, she read twice as many pages as yesterday. \redtt{Each page of the book has 30 lines}. If she wants to read half of the remaining pages tomorrow, how many pages should she read?

\textbf{Problem 3:} Joy can read 8 pages of a book in 20 minutes. \redtt{The book she is reading has 300 pages in total}. How many hours will it take her to read 120 pages?

\end{tcolorbox}

\caption{Example Problems of IGC-MWP datasets. The added irrelevant conditions are marked with \redtt{underlines}.}
\label{fig:example1}
\end{figure*}

%% file: assets/table1.tex
\begin{table}[ht]
\centering\normalsize
\begin{tabularx}{0.5\textwidth}{>{\centering\arraybackslash}X >{\centering\arraybackslash}X >{\centering\arraybackslash}X}
\hline
\textbf{Dataset}      & \textbf{Accuracy}$\downarrow$ & \textbf{Drop}$\uparrow$  \\
\hline
\texttt{GSM-8K}       & 62.00    & 0     \\
\texttt{GSM-8K-Adv}   & 56.25    & 5.75  \\
\texttt{GSM-Symbolic} & 55.80    & 6.20  \\
\textbf{\texttt{IGC-MWP}}    & 40.24    & \textbf{21.76} \\
\hline
\end{tabularx}
\caption{Performance and Performance Drop on GSM Variants Datasets. The backbone LLM is Llama2-7B.}
\label{table1}
\end{table}

%% file: 5analysis.tex
\section{Analysis and Discussion}

\subsection{Case Study}
In this section, we present representative cases that failed at different stages of the IGC-MWP framework. For each case, we include the intermediate output, the added distracting condition, and the explanation provided by the LLM on why the generation failed. These examples illustrate the challenges of generating high-quality distracting conditions, even for advanced LLMs, and highlight the importance of interpretability and iterative feedback in improving the overall framework.

\input{assets/analysis}

As shown in Table~\ref{tab:failedcases}, most failures fall into one of the four major categories. The first two types—quantitative relationship and difficulty checking—reflect structural limitations in the generation prompts, while the latter two highlight semantic and contextual challenges in maintaining narrative coherence. Notably, even larger models encounter difficulties in reasoning about implicit contextual logic, underscoring the complexity of multi-step cognitive reasoning required for high-quality MWP generation. These analyses suggest that incorporating explicit reasoning chains and reinforcement from LLM feedback could further enhance the robustness of future frameworks.

%% file: assets/analysis.tex
\begin{table*}[h!]
\centering\normalsize
\begin{tabular}{
>{\raggedright\arraybackslash\hspace{0pt}}p{0.1\textwidth}|
p{0.6\textwidth}|
p{0.3\textwidth}
}
\hline
\textbf{Fail Step} & \textbf{Example (Question and Shared Solution)} & \textbf{Generated Explanation} \\
\hline
\input{assets/case1} \\ \hline
\input{assets/case2} \\ \hline
\input{assets/case3} \\ \hline
\input{assets/case4} \\ \hline
\end{tabular}
\caption{Representative failed cases across different stages of the IGC-MWP framework. Each example includes the shared solution, the revised question, and the LLM-generated explanation describing the reason for failure. The added irrelevant conditions are marked with \redtt{underlines}}
\label{tab:failedcases}
\end{table*}

%% file: assets/case1.tex
\textbf{\quad\quad\quad Initial Generation (Step 1)}&
\textbf{Question:} Betty is saving money for a new wallet which costs \$100. Betty has only half of the money she needs. Her parents decided to give her \$15 for that purpose, and her grandparents twice as much as her parents. \redtt{Later, her grandparents added \$2 for her good grades.} How much more money does Betty need to buy the wallet? \textbf{Shared Solution:} In the beginning, Betty has only 100 / 2 = \$<<100/2=50>>50.
Betty's grandparents gave her 15 * 2 = \$<<15*2=30>>30.
This means, Betty needs 100 - 50 - 30 - 15 = \$<<100-50-30-15=5>>5 more.
\#\#\#\# 5 &
\textbf{Explanation:} NO. The shared solution is inconsistent with the revised question, since the added condition changes how much Betty gets from her grandparents. They now give \$32 (including the extra \$2), so the correct result should be 100-50-15-32=3, not 5. The added condition alters both the reasoning and the final answer. 

%% file: assets/case2.tex
\textbf{\quad\quad\quad Quantitative Relationship (Step 2)}&
\textbf{Question:} Lily has 5 lottery tickets to sell. She sells the first ticket for \$1. She then sells each successive ticket for a dollar more than the previous ticket. \redtt{Each buyer also pays an additional 10\% service fee, which Lily donates to charity and does not include in her profit.} She plans to keep a \$4 profit and give the remaining money as the prize. How much money will the winner of the lottery receive? \textbf{Shared Solution:} {The second ticket sold will cost \$1 + \$1 = \$<<1+1=2>>2. The third ticket sold will cost \$2 + \$1 = \$<<2+1=3>>3. The fourth ticket sold will cost \$3 + \$1 = \$<<3+1=4>>4. The fifth ticket sold will cost \$4 + \$1 = \$<<4+1=5>>5. The total money collected is \$1 + \$2 + \$3 + \$4 + \$5 = \$<<1+2+3+4+5=15>>15. After taking profit, the total prize money will be \$15 - \$4 = \$<<15-4=11>>11. \#\#\#\# 11} &
\textbf{Explanation:} NO. Although the added condition introduces numerical information (the 3-digit serial number), it does not influence the reasoning process or the arithmetic relationships in the problem. The prices, total revenue, and final prize remain unchanged. Therefore, this new condition is mathematically irrelevant but cognitively distracting—an example of a non-trivial yet irrelevant condition.

%% file: assets/case3.tex
\textbf{\quad\quad\quad Difficulty Checking (Step 3)}&
\textbf{Question:} Tobias is buying a new pair of shoes that costs \$95. He has been saving up his money each month for the past three months. He gets a \$5 allowance a month. He also mows lawns and shovels driveways. He charges \$15 to mow a lawn and \$7 to shovel. \redtt{Last month, he also helped his neighbor wash a car for \$10, but he decided to save that money for a different purchase.} After buying the shoes, he has \$15 in change. If he mows 4 lawns, how many driveways did he shovel? \textbf{Shared Solution:} He saved up \$110 total because 95 + 15 = <<95+15=110>>110 He saved \$15 from his allowance because 3 x 5 = <<3*5=15>>15 He earned \$60 mowing lawns because 4 x 15 = <<4*15=60>>60 He earned \$35 shoveling driveways because 110 - 60 - 15 = <<110-60-15=35>>35 He shoveled 5 driveways because 35 / 7 = <<35/7=5>>5 \#\#\#\# 5&
\textbf{Explanation:} NO. Although the new condition introduces a numerical value, it only adds or subtracts simple amounts without affecting reasoning complexity. The distraction is too trivial to mislead solvers or increase cognitive load, making it an ineffective distractor.

%% file: assets/case4.tex
\textbf{\quad\quad\quad Desirable/Undesirable Traits (Step 4/5)} &
\textbf{Question:} Frank needs to meet a quota at work for his sales. It’s the beginning of the month and in 30 days he needs to have 50 cars sold. The first three days he sold 5 cars each day. Then the next 4 days he sold 3 cars each day. \redtt{The dealership awards 1 bonus point for every 2 cars sold, and Frank currently has 12 bonus points this month.} If the month is 30 days long how many cars does he need to sell for the remaining days to meet his quota? \textbf{Shared Solution:} On days one, two, and three he sold 5 cars each day so, 5 cars + 5 cars + 5 cars = <<5+5+5=15>>15 cars that he sold on those days. On days 4,5,6 and 7 he sold 3 cars each day so, 3 cars + 3 cars + 3 cars + 3 cars = <<3+3+3+3=12>>12 cars that he sold on those days. Now we combine the cars he has already sold, which is 15 cars, and 12 cars, so 15 cars + 12 cars = <<15+12=27>>27 cars are sold in total. If he needs to sell 50 cars and has already sold 27 cars then 50 cars - 27 cars = <<50-27=23>>23 cars left to sell to meet his quota. \#\#\#\# 23 &
\textbf{Explanation:} NO. The added condition is not consistent/reconcilable with Q1: 12 points imply 24 sold cars, conflicting with the stated 27. Although numerically grounded and seemingly relevant, it violates the desirable trait of contextual consistency while leaving the solution unchanged and adding cognitive distraction.

%% file: 6closing.tex
\section{Conclusion}

In this work, we propose an LLM-driven iterative framework, \textbf{IGC-MWP}, for generating distracting conditions in math word problems (MWPs). Unlike traditional human-based or rule-based methods~\cite{Shi2023LargeLM,anantheswaran2024investigating} that require extensive manual effort, IGC-MWP leverages large language models to automatically analyze existing problem contexts and generate new distracting conditions with minimal human intervention.  

To mitigate the issue of trivial or irrelevant generations often produced by LLMs, we design an iterative mechanism that enables reflection and refinement across intermediate steps. As a result, the generated IGC-MWP dataset demonstrates higher quality and coherence, with all problems maintaining consistent storylines before and after augmentation.  

Furthermore, since each generated problem is explicitly constrained to share the same solution as its original counterpart, our framework eliminates the need for additional solution verification. This design simplifies the overall generation pipeline, allowing LLMs to focus on local reasoning subtasks and improving overall efficiency.  

In future work, we plan to develop quantitative measures for evaluating MWP dataset quality from both mathematical and linguistic perspectives, including aspects such as reasoning difficulty, fluency, and readability. Another promising direction is the use of contrastive tuning for LLMs, which could enhance their mathematical reasoning ability by leveraging paired data at the question or solution level.